\documentclass[10pt,twocolumn,letterpaper]{article}
\pdfoutput=1
\usepackage{cvpr}
\usepackage{times}
\usepackage{epsfig}
\usepackage{graphicx}
\usepackage{amsmath}
\usepackage{amssymb}

\usepackage{array}
\usepackage{multirow}
\usepackage{subfig}

\DeclareCaptionFormat{myformat}{#1#2#3\hrulefill}
\usepackage[pagebackref=true,breaklinks=true,colorlinks,bookmarks=false]{hyperref}
\cvprfinalcopy 

\def\cvprPaperID{6328} 

\ifcvprfinal\fi

\begin{document}

\title{SceneCode: Monocular Dense Semantic Reconstruction using Learned Encoded Scene Representations}
\author{Shuaifeng Zhi, Michael Bloesch, Stefan Leutenegger, Andrew J. Davison\\
Dyson Robotics Laboratory at Imperial College\\
Department of Computing, Imperial College London, UK\\
{\tt\small \{s.zhi17, m.bloesch, s.leutenegger, a.davison\}@imperial.ac.uk}
}

\maketitle

\begin{abstract}
Systems which incrementally create 3D semantic maps from image sequences must store and update representations of both geometry and semantic entities. However, while there has been much work on the correct formulation for geometrical estimation, state-of-the-art systems usually rely on simple semantic representations which store and update independent label estimates for each surface element (depth pixels, surfels, or voxels).
Spatial correlation is discarded, and fused label maps are incoherent and noisy.

We introduce a new compact and optimisable semantic representation by training a variational auto-encoder that is conditioned on a colour image.
Using this learned latent space, we can tackle semantic label fusion 
by jointly optimising the low-dimenional codes associated with each of a set of overlapping images, producing consistent fused label maps
which preserve spatial correlation.
We also show how this approach can be used within a monocular keyframe based semantic mapping system where a similar code approach is used for geometry. The probabilistic formulation allows a flexible formulation where we can jointly estimate motion, geometry and semantics in a unified optimisation.
\end{abstract}
\section{Introduction}

Intelligent embodied devices such as robots need to build and maintain
representations of their environments which permit inference of geometric and semantic properties, such as the traversability 
of rooms or the way to grasp objects. Crucially, if this inference is
to be scalable in terms of computing resources, these representations
must be {\em efficient}; and if devices are to operate {\em robustly},
the employed representations must cope with all of the variation present in
the real world. However, current real-time scene understanding systems are
still a long way from the performance needed for truly ground-breaking
applications \cite{Cadena:etal:TRO2016,Davison:ARXIV2018}.
\begin{figure}[t]
\centering
\includegraphics[width=0.9\columnwidth]{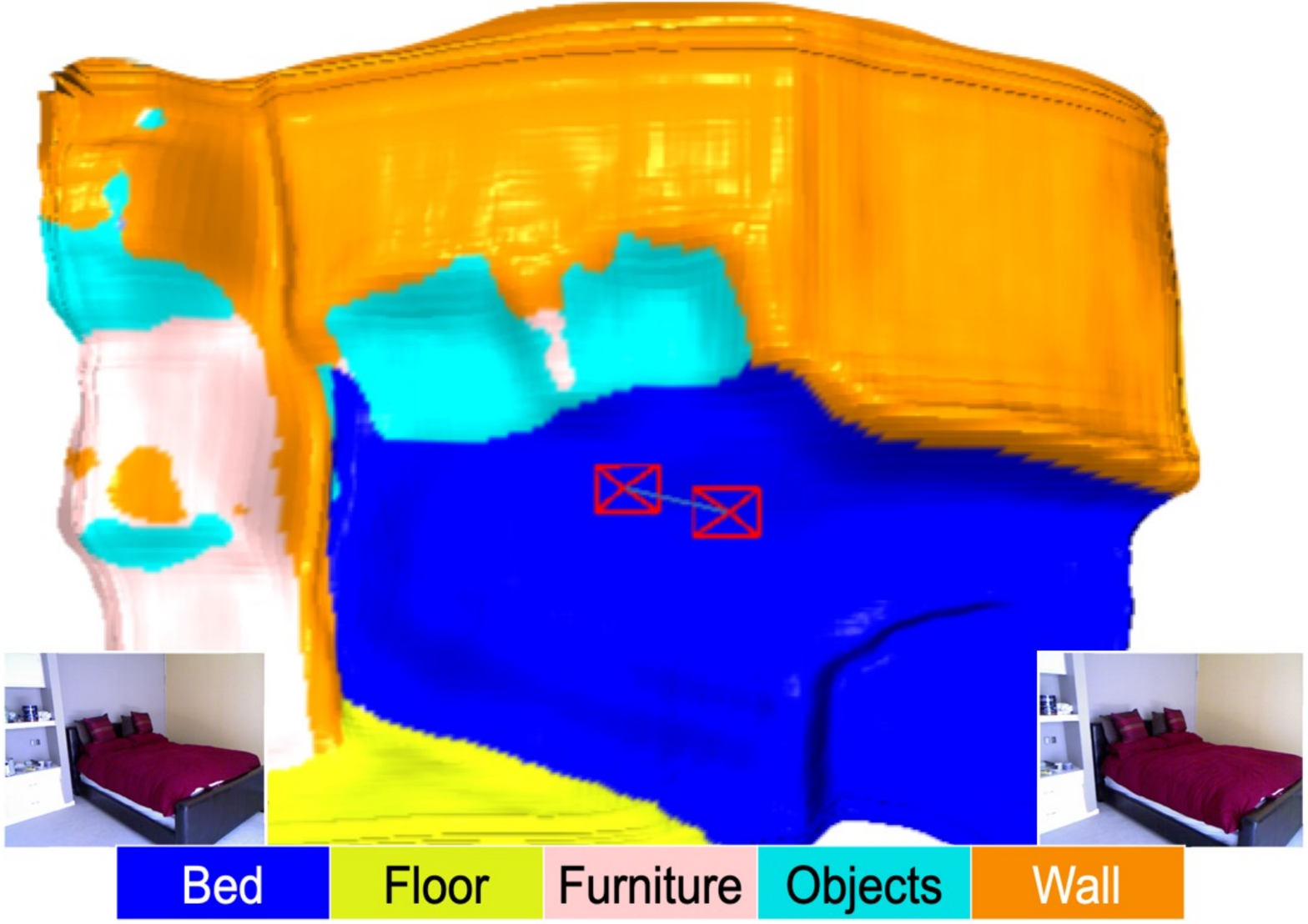}
\caption{Dense semantic structure from motion on two frames from the NYUv2 dataset. Compact representations of semantics and geometry have been jointly optimised with camera pose to obtain smooth and consistent estimates. \label{fig:2view_SFM_bedroom}}
\end{figure}

An eventual token-like, composable scene understanding may finally
give artificial systems the capability to reason about space and shape
in the intuitive manner of humans~\cite{Jiajun:etal:NIPS2015}.
Bringing deep learning into traditional hand-designed estimation
methods for SLAM has certainly led to big advances to representations
which can capture both shape and semantics \cite{Cadena:etal:RSS2016,Weerasekera:etal:ICRA2017}, but so far these are
problematic in various ways. The most straightforward approaches, such
as~\cite{Kahler:Reid:ICCV2013,Hermans:etal:ICRA2014,McCormac:etal:ICRA2017,Xiang:Fox:ARXIV2017,Xiao:Quan:ICCV2009,Ma:etal:IROS2017} which paint a
dense geometric SLAM map with fused semantic labels predicted from
views~\cite{Long:etal:CVPR2015, Lin:etal:CVPR2017, He:etal:CVPR2017}, are expensive in
terms of representation size; label scenes in an
incoherent way where each surface element independently stores its
class; and do not benefit from semantic labelling improving motion or geometry estimation.

At the other end of the scale are approaches which explicitly
recognise object instances and build scene models as 3D object
graphs~\cite{Mccormac:etal:3DV2018,Sunderhauf:etal:IROS2017,Nicholson:etal:RAL2018, Runz:etal:ISMAR2018}. These representations have the
token-like character we are looking for, but are limited to
mapping discrete `blob-like' objects from known classes and leave
large fractions of scenes undescribed.

Looking for efficient representations of whole scenes, in this work we
are inspired by CodeSLAM from Bloesch \etal~\cite{Bloesch:etal:CVPR2018} which used a learned encoding to represent the dense geometry of a scene with small codes which can be
efficiently stored and jointly optimised in multi-view SLAM.
While~\cite{Bloesch:etal:CVPR2018} encoded only geometry, here we show
that we can extend the same conditional variational auto-encoder
(CVAE) to represent the multimodal distributions of semantic segmentation.
As in CodeSLAM, our learned low-dimensional semantic code
is especially suitable for, but not limited to keyframe based semantic
mapping systems, and allows for joint optimisation across multiple
views to maximise semantic consistency. This joint optimisation 
alleviates the problems caused by the independence of surface elements
assumed by most semantic fusion methods, and allows much higher
quality multi-view labellings which preserve whole elements of natural
scenes. 

We show that compact representations of geometry and semantics can be
jointly learned, resulting in the multitask CVAE used in this
paper. This network makes it possible to build a monocular dense
semantic SLAM system where geometry, poses and semantics can be jointly
optimised.

To summarise, our paper has the following contributions:

\begin{itemize}
\item A compact and optimisable representation of semantic segmentation using an image-conditioned variational auto-encoder.

\item A new multi-view semantic label fusion method optimising semantic consistency.

\item A monocular dense semantic 3D reconstruction system, where geometry and semantics are tightly coupled into a joint optimisation framework.

\end{itemize}

\section{Related Work} \label{sec:Related Work}

Structured semantic segmentations of the type we propose
have been studied by several authors. Sohn \etal~\cite{Sohn:etal:NIPS2015}
proposed a CVAE to learn the
distribution of object segmentation labels using Gaussian latent
variables. Due to the learned distribution, the resulting object
segmentation was more robust to noisy and partially observed data
compared to discriminative CNN models. Pix2Pix from Isola \etal~\cite{Isola:etal:CVPR2017} used a conditional Generative
Adversarial Network (GAN) to achieve image to image translation tasks
in which the conditional distribution of semantic labels is implicitly
learned. However, when used for semantic prediction from colour
images, the GAN training process induces hallucinated objects.
In addition, the distributions learned by GANs are not directly
accessible and optimisable in the form we need for multi-view fusion.

Kohl \etal recently proposed a probabilistic U-Net
\cite{Kohl:etal:NIPS2018} to address the ambiguities of semantic
segmentation due to insufficient context information.  A CVAE was designed to learn the multimodal distribution of segmentations
given colour images through a low-dimensional latent space, and it was shown that ambiguities can be well modelled by a compact latent code. 
We build on this idea and show that we can use  the learned latent space to integrate multi-view semantic labels, and build a monocular dense SLAM system
capable of jointly optimising geometry and semantics.

\begin{figure*}[!ht]
\centering
\includegraphics[height=6.8cm,width=1.9\columnwidth]{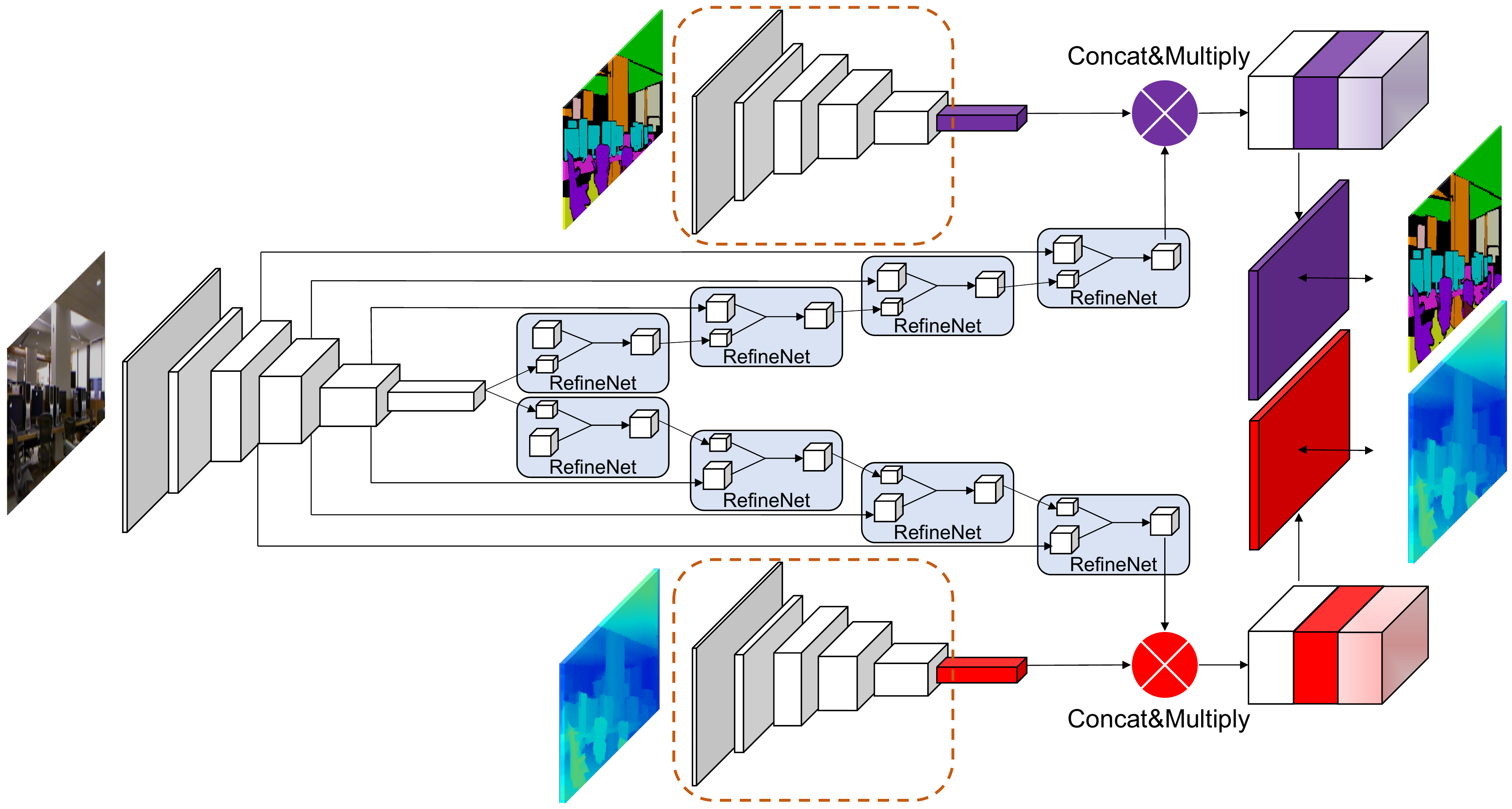}
\caption{The proposed multitask conditional variational auto-encoder (multitask CVAE). Depth images and semantic labels (one-hot encoded) are encoded to two low-dimensional latent codes via VGG-like fully convolutional networks. These recognition models shown in the dashed boxes are not accessible during inference. The RGB images are processed by a U-shaped
network with a ResNet-50 backbone. Finally, the sub-parts are combined by $\otimes$ operations standing for a combination of broadcasting, concatenation, and element-wise multiplication.\label{fig:multitask CVAE}}
\end{figure*}
\section{Compact Geometry + Semantics Encoding} \label{sec: Compact Map Representation}
Our multitask CVAE (see Figure \ref{fig:multitask
  CVAE}) learns the conditional probability densities for
depths and semantic segmentations conditioned on colour images
in a manner similar to the compact representation of geometry in
CodeSLAM~\cite{Bloesch:etal:CVPR2018}.
The network consists of three main parts: a U-shaped multitask network with skip connections and two variational auto-encoders for depth and semantic segmentation.

The U-shaped multitask network contains one shared encoder with a
ResNet-50 backbone \cite{He:etal:CVPR2016} and two separate decoders
adopting RefineNet units \cite{Lin:etal:CVPR2017}. 
Unlike the original implementation, batch normalisation is added after each convolution in the RefineNet unit to stabilise training. Each of the two variational auto-encoders consists of a VGG-like fully convolutional
recognition model (encoder) followed by a linear generative model
(decoder), which is coupled with the U-net and thus conditioned on colour images.

More specifically, in the linear decoder the latent code is first
broadcast spatially to have the same height/width and then \textbf{$1\times1$}
convolved to have the same dimensionality as the image feature maps
from the last RefineNet unit.
A merged tensor is then computed by doing a three-fold concatenation of the broadcast/convolved code, the RefineNet unit, and an element-wise multiplication of the two.
Finally, convolution (without nonlinear activation) and bilinear upsampling is applied to obtain the prediction.
The motivation for this procedure is to obtain a linear relationship between code and prediction which is conditioned on the input image in a nonlinear manner \cite{Bloesch:etal:CVPR2018} --- the linearity enabling pre-computation of Jacobians during inference at test time (see Section \ref{sec:Dense Warping}).
The predicted depth and semantics (unscaled logits before softmax function) can thus be formulated as:
\begin{gather}
D\left(\boldsymbol{c}_d,I\right) = D_0\left(I\right) + J_d\left(I\right)\boldsymbol{c}_d,\label{eq:depth_formulation}\\
S\left(\boldsymbol{c}_s,I\right) = S_0\left(I\right) + J_s\left(I\right)\boldsymbol{c}_s,\label{eq:semantic_formulation}
\end{gather}
where $J_{s/d}$ represents
the learned linear influence, and $D_0(I) = D(0, I)$ and $S_0(I) = S(0, I)$.
Due to our variational setup, $D_0(I)$ and $S_0(I)$ can be interpreted as the most likely prediction given the input image alone.
Note the generality of this framework, which could be combined with arbitrary network architectures.

\subsection{Network Training Configuration}\label{sec:training}
Both the depth and semantic predictions are jointly trained using groundtruth data.
In addition to the reconstruction losses discussed in the following sections, the variational setup requires a KL-divergence based loss on the latent space \cite{Kingma:Welling:ICLR2014}.
In order to avoid a degrading latent space, we employ a KL annealing strategy \cite{Bowman:etal:ARVIX2015,Sonderby:etal:arXiv2016} where we gradually increase the weights of the KL terms from 0 after 2 training epochs.
Finally, the weights of semantic vs.\ depth reconstruction losses are trained in an adaptive manner to account for task-dependent uncertainty \cite{Kendall:etal:CVPR2018}.
In all of our experiments, we train
the whole network in an end-to-end manner using the Adam optimiser \cite{Kingma:Ba:ICLR2015} with an initial learning rate of $10^{-4}$ and a weight decay of
$10^{-4}$. The ResNet-50 is initialised using ImageNet pre-trained
weights, and all other weights are initialised using He \etal's method  \cite{He:etal:ICCV2015}.

For depth images, as in \cite{Bloesch:etal:CVPR2018},
the raw depth values $d$ are first transformed via a\emph{ }hybrid
parametrisation\emph{ }called\emph{ proximity}, $p= a/(a+d)$,
where $a$ is the average depth value, which is set to 2m in all
of our experiments. In this way, we can handle raw depth values
ranging from $0$ to $+\infty$ and assign more precision
to regions closer to the camera. An $L_{1}$ loss function
together with data dependent Homoscedastic uncertainty \cite{Kendall:Gal:NIPS2017}
is used as the reconstruction error:
\begin{align}
L_{\phi,\theta}\left(d\right) & = \sum_{i=1}^{N}\left[\frac{\left|\widetilde{p}_{i}-p_{i}\right|}{b_{i}}+\log\left(b_{i}\right)\right],\label{eq:DepthCVAE_loss}
\end{align}
where $N$ is the number of pixels, $\widetilde{p}_{i}$
and $p_{i}$ are the predicted proximity and input proximity of
the $i$-th pixel, and $b_{i}$ is the predicted uncertainty of the $i$th
pixel.

Semantic segmentation labels, which are
discrete numbers, are one-hot encoded before being input to
the network. Therefore, the multi-class cross-entropy function is
a natural option for calculating the reconstruction loss using the predicted
softmax probabilities and one-hot encoded labels:
\begin{align}
L_{\phi,\theta}\left(s\right) & = \frac{1}{N}\sum_{i=1}^{N}\sum_{c=1}^{C}k_{c}^{\left(i\right)}\log p_{c}^{\left(i\right)},\label{eq:SemanticCVAE_loss}
\end{align}
where $C$ is the number
of classes, $k_{c}^{\left(i\right)}$ is the $c$-th element of the one-hot
encoded labels for the $i$-th pixel and $p_{c}^{\left(i\right)}$ is the
predicted softmax probability in the same position.

\section{Fusion via Multi-View Code Optimisation}\label{sec:Dense Warping}
In a multi-view setup, depth, semantics, and motion estimates can be refined based on consistency in overlapping regions by making use of dense correspondences.
While the use of photometric consistency is well-established, here we also introduce semantic consistency, i.e.\ any given part of our scene should have the same semantic label irrespective of viewpoint.
The semantic consistency is less affected by disturbances such as non-Lambertian reflectance, but may be subject to quantisation errors and cannot be directly measured.

Given no additional information, an all-zero code is most likely code because of the multivariate Gaussian prior assumption during training (Section \ref{sec:training}). This zero code can thus be used, both as an initialisation value and as a prior during optimisation at test time (during which we have no access to depths or semantic labels). The probabilistic formulation of the system allows it to embed depth, semantics and motion into a unified probabilistic framework and thereby combine an arbitrary number of information sources including images, semantic constraints, priors, motion models or even measurements from other sensors.

\subsection{Geometry Refinement}

In analogy to \cite{Bloesch:etal:CVPR2018}, given
an image $I_{A}$ with its depth code $\boldsymbol{c}_d^A$, and a
second image $I_{B}$ with estimated relative rigid body transformation $\boldsymbol{T}_{BA}=\left(\boldsymbol{R}_{BA},\boldsymbol{t}_{BA}\right)\in SO\left(3\right)\times\mathbb{R}^{3}$, the dense correspondence for each pixel $\boldsymbol{u}$ in view A is:
\begin{equation}
w\left(\boldsymbol{u}_{A},\boldsymbol{c}_d^A,\boldsymbol{T}_{BA}\right)=\pi\left(\boldsymbol{T}_{BA} \, \pi^{-1}\left(\boldsymbol{u}_{A},D_{A}\left[\boldsymbol{u}_{A}\right]\right)\right),
\end{equation}
where $\pi$ and $\pi^{-1}$ are the projection and inverse projection functions, respectively. $D_{A}$ stands for $D_{A}=D\left(\boldsymbol{c}_d^A,I_{A}\right)$,
and the square bracket operation $\left[\boldsymbol{u}\right]$ means a
value look-up at pixel location $\boldsymbol{u}$. We can then establish the photometric error $\mathit{r}_{i}$ based on the photo-consistency assumption \cite{Kerl:etal:ICRA2013}:
\begin{equation}
\mathit{r}_{i}=I_{A}\left[\boldsymbol{u}_{A}\right]-I_{B}\left[w\left(\boldsymbol{u}_{A},\boldsymbol{c}_d^A,\boldsymbol{T}_{BA}\right)\right].\label{eq:photometric_error}
\end{equation}
Similarly, we can derive the geometric error term $\mathit{r}_{z}$ as:
\begin{equation}
\mathit{r}_{z}=D_{B}[w(\boldsymbol{u}_{A},\boldsymbol{c}_d^A,\boldsymbol{T}_{BA})]-[\boldsymbol{T}_{BA}\pi^{-1}(\boldsymbol{u}_{A},D_{A}[\boldsymbol{u}_{A}])]_{\mathit{Z}}\label{eq:geometric_error},
\end{equation}
where $\left[\cdot\right]_{\mathit{Z}}$ refers to the depth value of a point.

Both photometric and geometric errors are differentiable w.r.t.\ the input camera poses and latent codes, so that Jacobians can be computed using the chain rule.
Due to the designed linear relationship we can pre-compute the Jacobian of depth prediction w.r.t.\ the code which is computationally expensive due to the dense convolution operations.

\subsection{Semantics Refinement}
Given images $I_{A}$ , $I_{B}$ sharing
a common field of view (FOV), and their pre-softmax predictions $S_{A}$ and $S_{B}$ generated
from semantic codes $\boldsymbol{c}_s^A$ and $\boldsymbol{c}_s^B$, we propose to establish a semantic error term via dense warping:
\begin{equation}
r_{s}=DS\left(S_{A}\left[\mathbf{u}_{A}\right],S_{B}\left[w\left(\mathbf{u_{A}},\boldsymbol{c}_d^A,\boldsymbol{T}_{BA}\right)\right]\right),\label{eq:semantric_error_vanilla}
\end{equation}
where $DS$ can be an arbitrary function measuring distance/dissimilarity \cite{Cha:Srihari:PR2002}.
In the scope of this paper, $DS$ is chosen to be the Euclidean distance
after applying softmax on the logits. Establishing the semantic error on top of semantic labels is not adopted here due to the loss of information and the induced non-differentiability. 

The underlying intuition of Equation \ref{eq:semantric_error_vanilla} 
is that corresponding pixels must have the same semantic label, and thus similar (but not necessary the same) softmax categorical probabilities.
Unlike the photo-consistency assumption, the semantic consistency assumption is comparatively weak since it is not anchored to any actual measurement, though this is somewhat alleviated by the zero-code prior described above.
Nevertheless, as the viewpoint varies, different semantic cues may become available and a previously semantically ambiguous region may become more distinctive.
Instead of fusing this information element-wise \cite{McCormac:etal:ICRA2017}, the estimates are propagated all the way back to the semantic code, allowing spatial information fusion.

The semantic error term in Equation \ref{eq:semantric_error_vanilla} is
differentiable not only w.r.t.\ the semantic codes $\boldsymbol{c}_s^A$
and $\boldsymbol{c}_s^B$, but also w.r.t.\ the camera pose and depth of the reference keyframe.
This naturally enables semantic information to influence motion and structure estimation, i.e., the framework will for instance attempt to align chairs with chairs and walls with walls.
Again, the Jacobians of the semantic logits w.r.t.\ the semantic code can be pre-computed.

Although our semantics refinement approach targets a monocular
keyframe based SLAM system, it can be adopted as a semantic
label fusion module in arbitrary SLAM system such as stereo or RGB-D SLAM systems.

\begin{figure}
\centering
\includegraphics[width = 0.9\columnwidth]{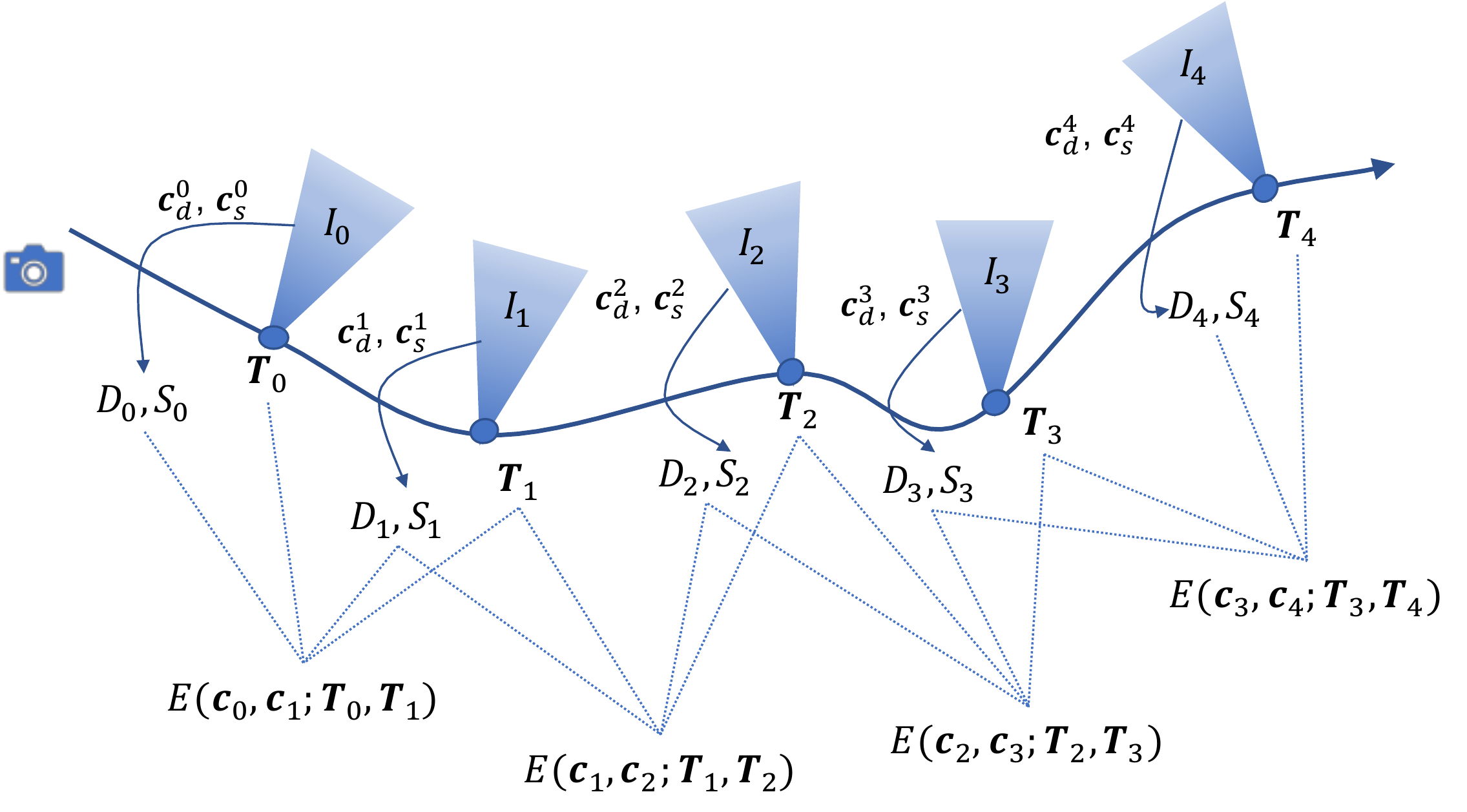}
\caption{Semantic mapping formulation. Each keyframe has a colour
image $I$, depth code $\boldsymbol{c}_d$ and semantic code $\boldsymbol{c}_s$. Second order
optimisation can be applied to jointly or separately optimise
camera motion, geometry and semantics. \label{fig:Illustration-of-SFM}}
\end{figure}

\section{Monocular Dense Semantic SLAM}
We can integrate the geometry and semantics refinement
processes into a preliminary keyframe based monocular SLAM system. The
map is represented by a collection of keyframes, each with
a camera pose and two latent codes, one for geometry
and one for semantics, as shown in Figure \ref{fig:Illustration-of-SFM}.
We follow the standard paradigm of dividing
the system into tracking (front-end) and mapping (back-end)
and alternate between them \cite{Klein:Murray:ISMAR2007}.
In the present paper, for efficiency reasons, the tracking module estimates the relative 3D motion between the current frame and the last keyframe using the photometric residual only \cite{Baker:Matthews:IJCV2004}.

The mapping module relies on dense N-frame structure from motion,
by minimising photometric, geometric and semantic residuals with a zero-code prior between any two overlapping frames, which can be formulated as a non-linear least-square problem.
As in CodeSLAM \cite{Bloesch:etal:CVPR2018}, we employ loss functions that (i) remove invalid
correspondences, (ii) perform relative weighting for different residuals,
(iii) include robust Huber weighting, (iv) down-weight strongly slanted
and potentially occluded pixels. The differentiable residuals are
minimised by a damped Gauss-Newton.
In addition, the linear decoder allows us to pre-compute the Jacobians of
the network prediction w.r.t.\ the code for each keyframe. Because the semantic residual relies not only on the semantic code but also on data association, during mapping we first jointly optimise the geometry and poses, then optimise the semantic residual, and lastly jointly optimise both geometry and semantics. In this way, we tightly couple geometry and semantics into a single optimisation framework.

\begin{figure}[!t]
\centering
\includegraphics[width =0.5\columnwidth]{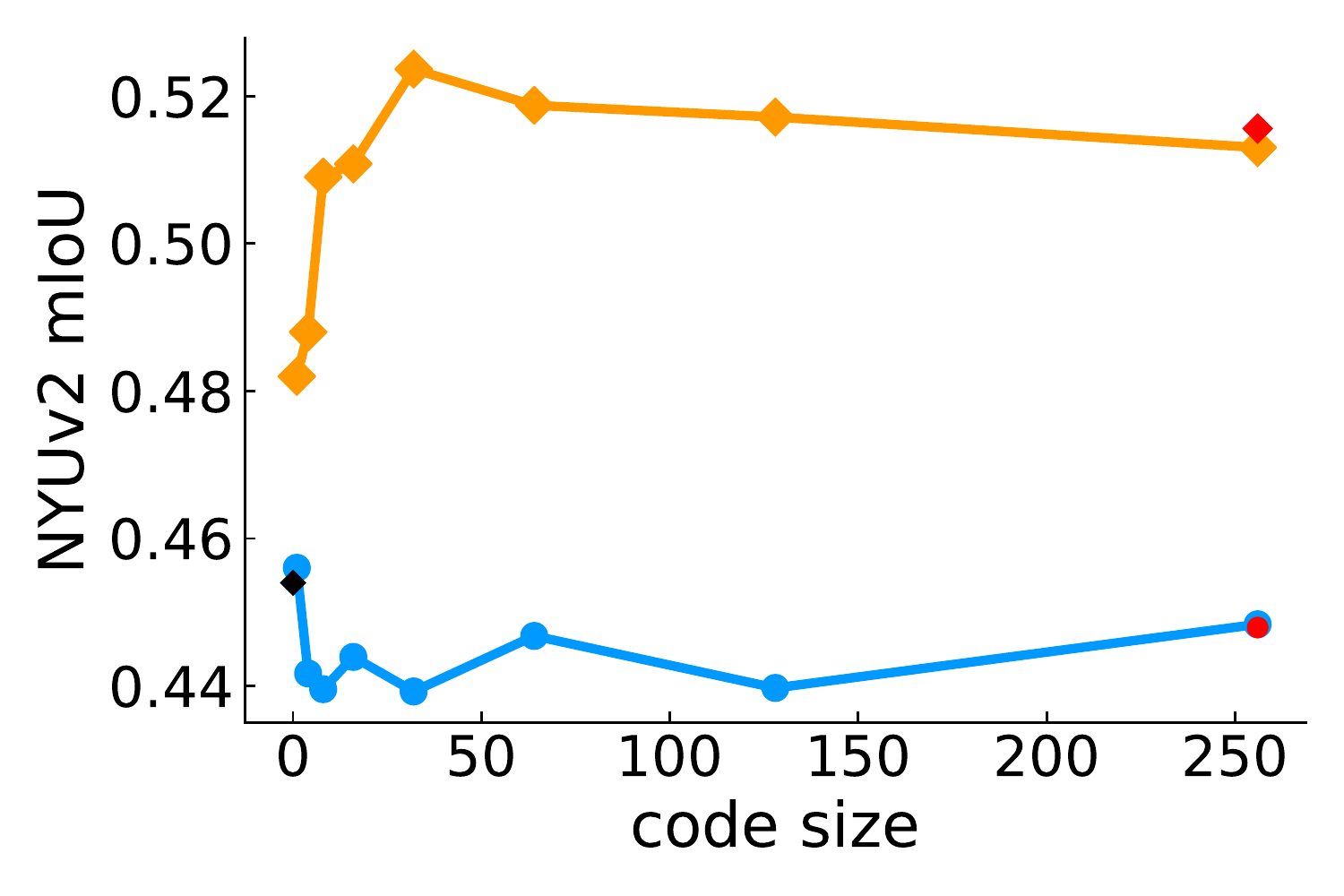}\includegraphics[width =0.5\columnwidth]{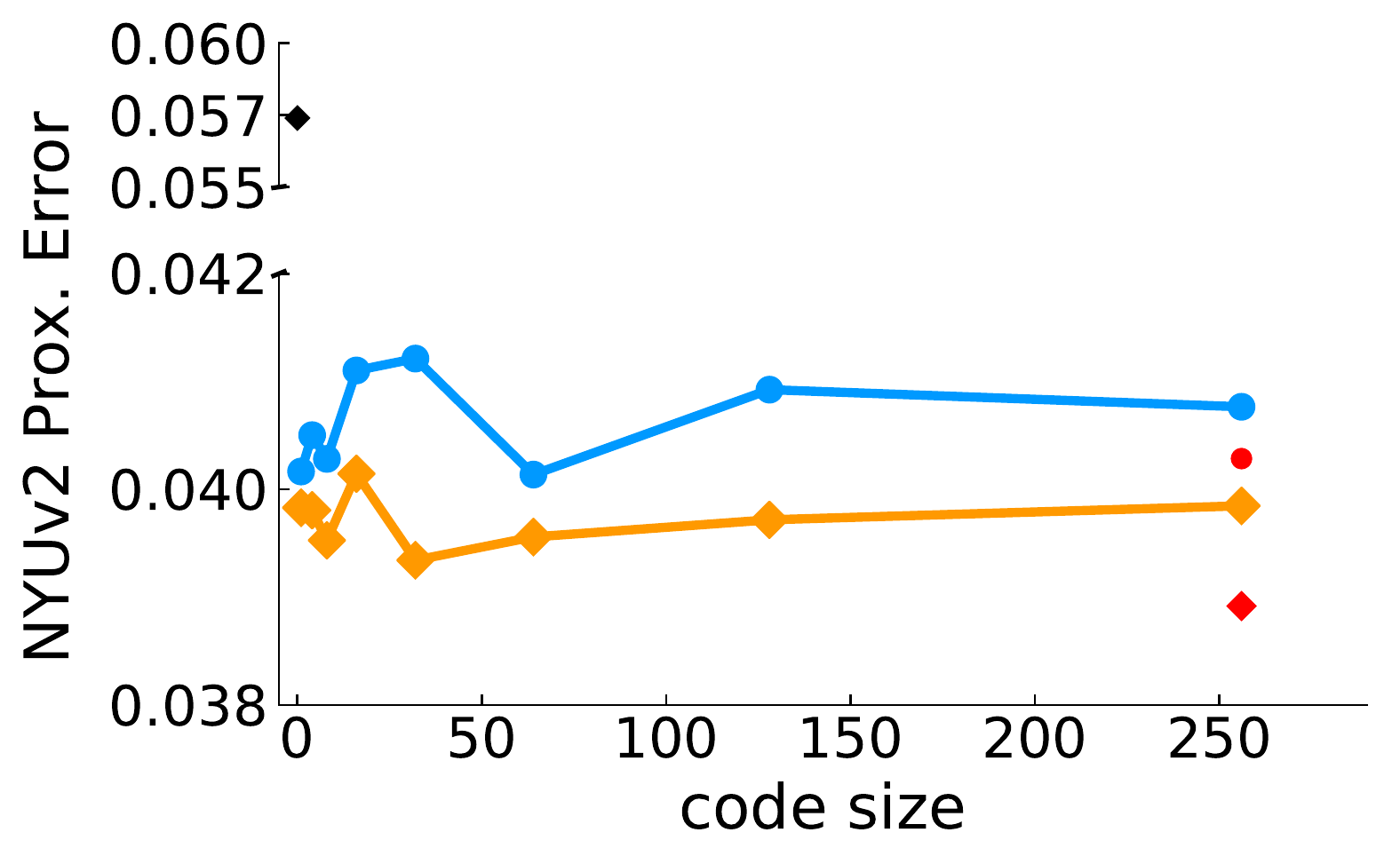}
\centering
\includegraphics[width =0.5\columnwidth]{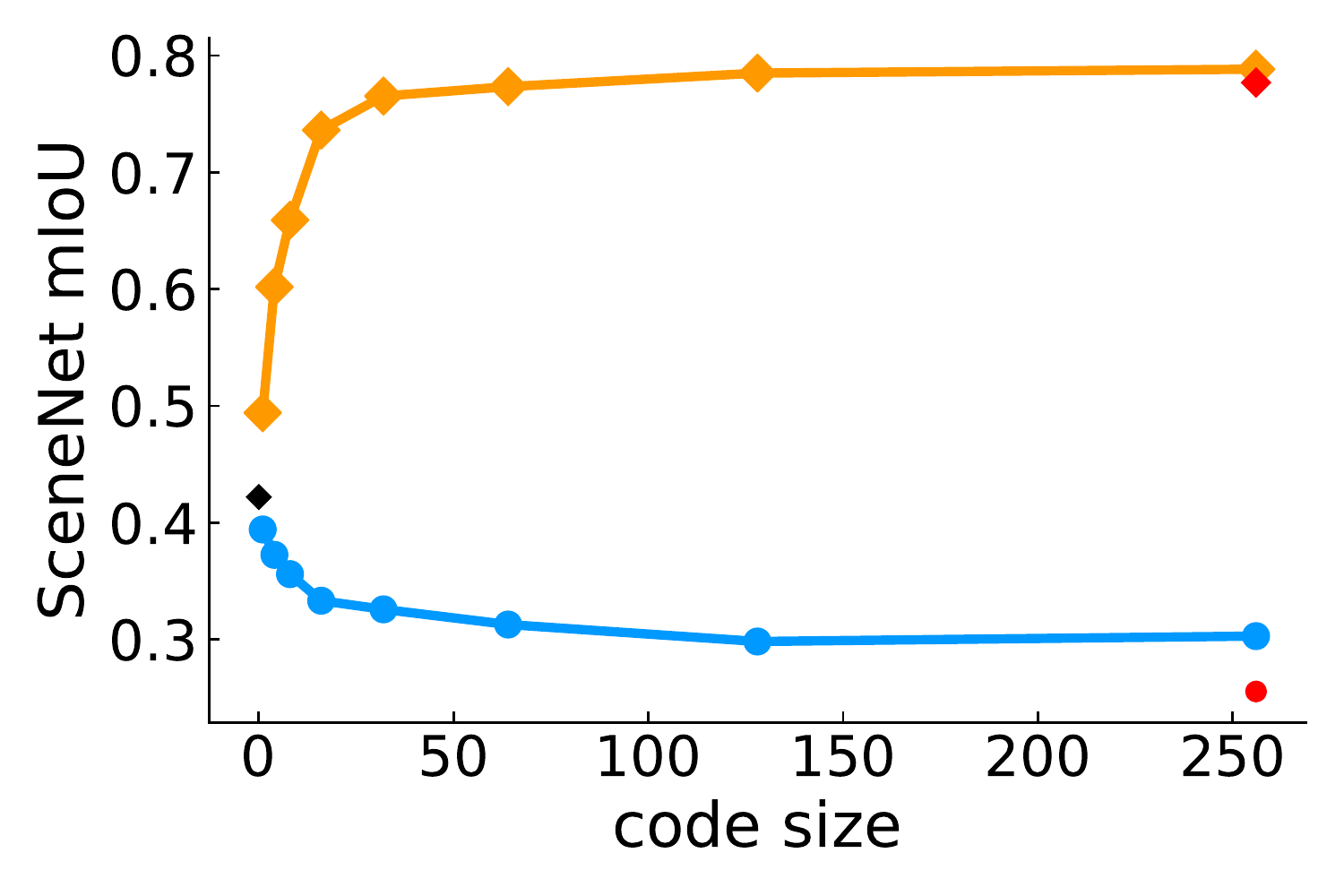}\includegraphics[width =0.5\columnwidth]{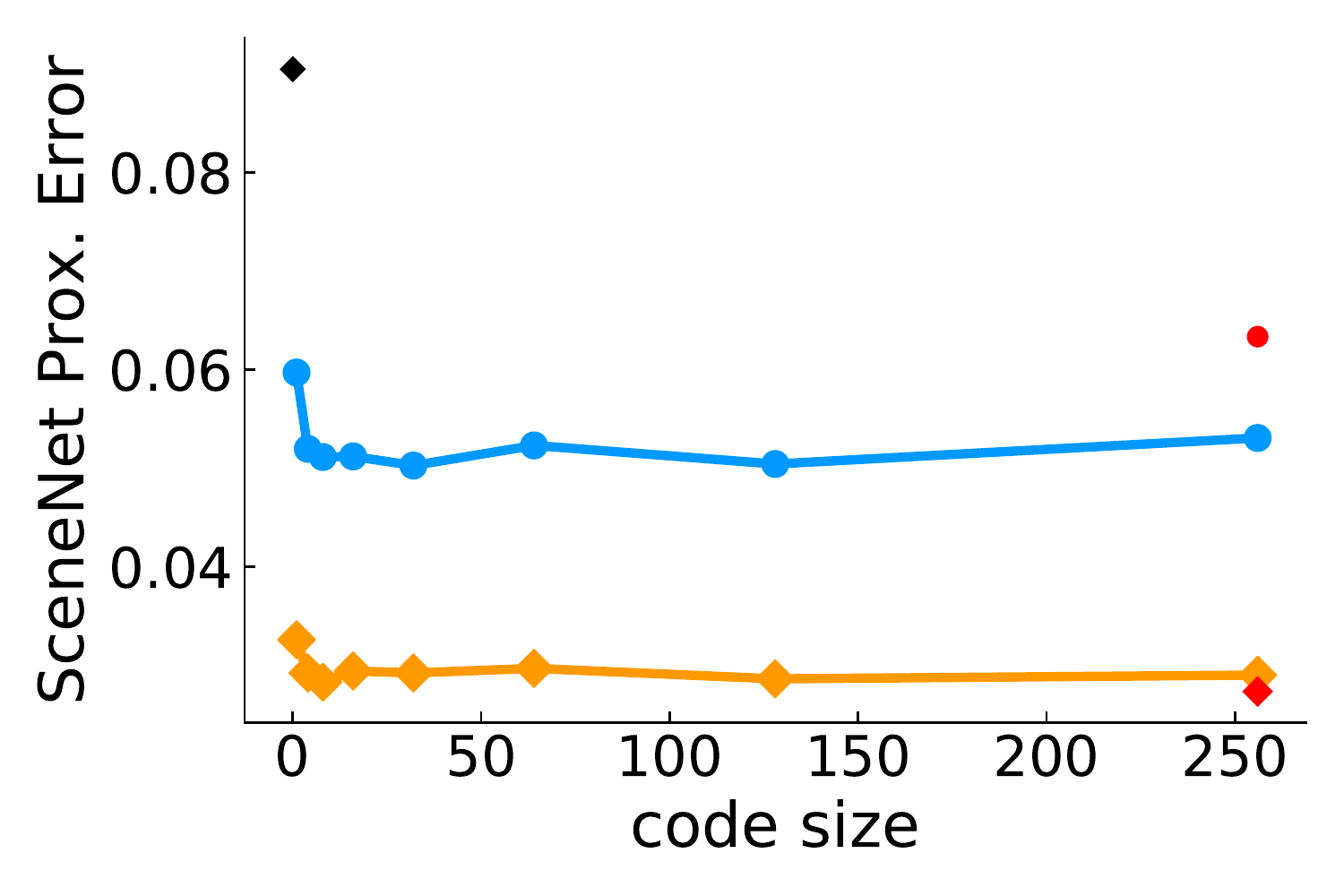}

\begin{minipage}[t]{\columnwidth}%
\centering
\includegraphics[width = 0.95\columnwidth]{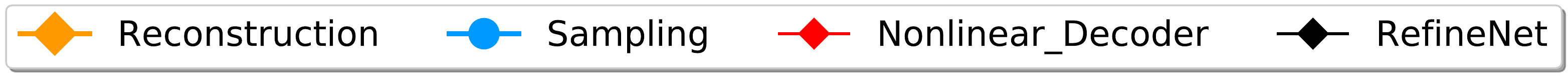}
\end{minipage}
\caption{Reconstruction and zero-code prediction performance of different set-ups on the NYUv2 and SceneNet RGB-D test sets. Reconstruction performance increases with code size. The quality of zero-code predictions is comparable to a discriminative RefineNet model for semantic segmentation,  and better on depth prediction. Using a non-linear decoder leads to little improvement.
\label{fig:The-reconstruction-accuracy}}
\end{figure}

\begin{figure*}[!t]
\centering
\includegraphics[height = 0.361\pdfpageheight, width=1.95\columnwidth]{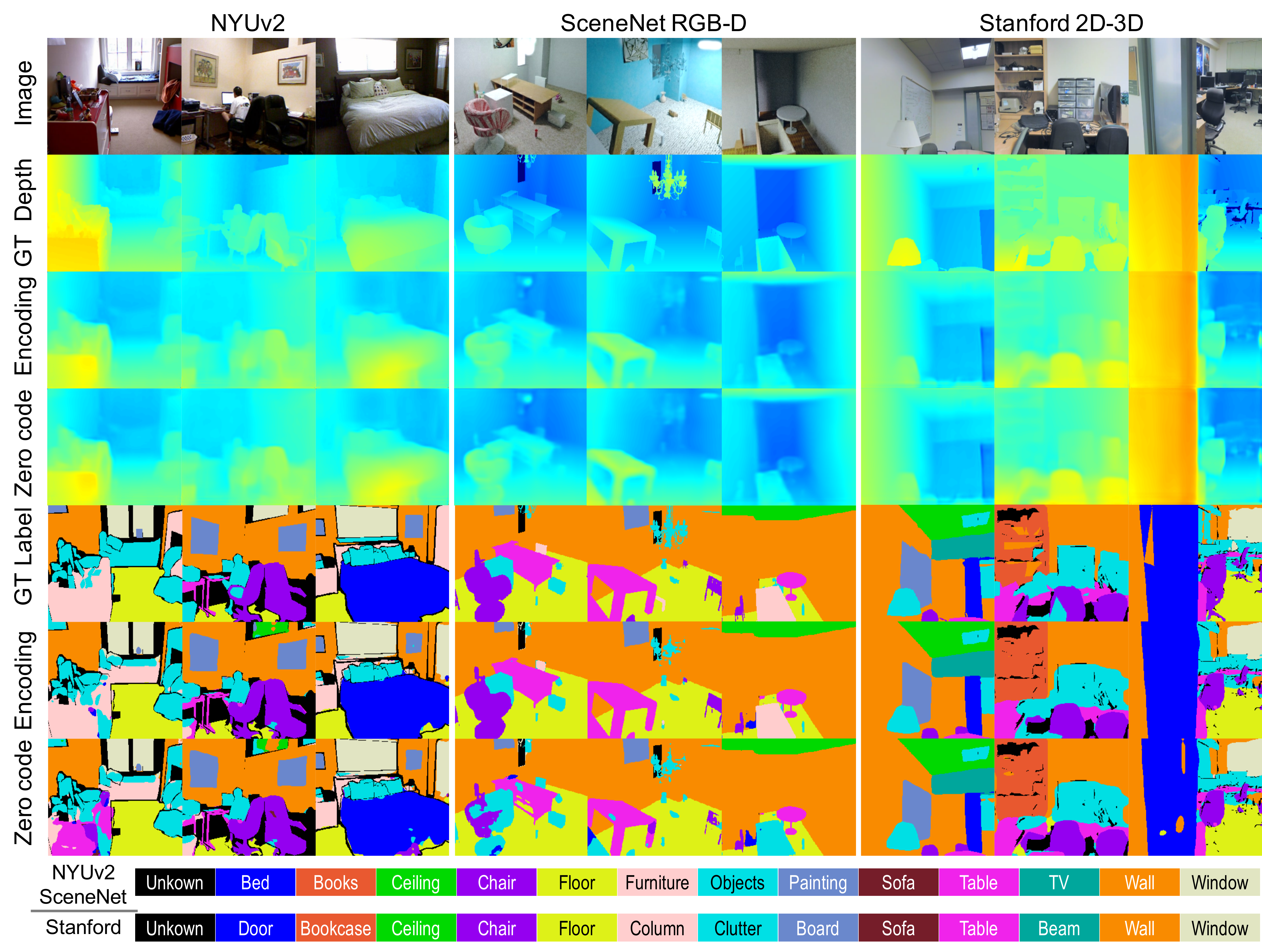}
\caption{Qualitative results on the NYUv2 (left), SceneNetRGB-D (middle) and Stanford (right) datasets. Input colour images are at the top. We show ground truth, encoded predictions (code from encoder) and zero-code predictions (monocular predictions) for depth and semantic labels. Incorrect semantic predictions in regions which are ambiguous for monocular predictions are corrected by optimising the compact latent codes. Black regions are masked unknown classes. \label{fig:Qualitative-results}}
\end{figure*}

\section{Experiments}
Please also see our submitted video which includes further demonstrations: \href{https://youtu.be/MCgbgW3WA1M}
{https://youtu.be/MCgbgW3WA1M}.

To test our method, we use three indoor datasets: the synthetic SceneNet RGB-D
\cite{McCormac:etal:ICCV2017} dataset, and the real-world NYUv2
\cite{Silberman:etal:ECCV2012} and Stanford 2D-3D-Semantic datasets
\cite{Armeni:etal:ARXIV2017}. 
Compared to outdoor road scenes
\cite{Geiger:etal:CVPR2012,Cordts:etal:CVPR2016}, indoor scenes have different challenges with large variations in
spatial arrangement and object sizes, and full 6-D motion. 

\subsection{Datasets}
\textbf{NYUv2} has 1,449 pre-aligned and annotated images (795 in
the training set and 654 in the test set). We cropped all the available images from 640 $\times$ 480 to valid regions of 560 $\times$ 425 before further
processing. The 13 class semantic segmentation task is evaluated in our experiments.

\textbf{Stanford 2D-3D-Semantic} is a large scale real world
dataset with a different set of 13 semantic class definitions. 70,496 images with random camera parameters are split into a training set of 66,792 images (areas 1, 2, 4, 5, 6) and a test set of 3,704 images (area 3). We rectified all images to a unified camera model.

The synthetic \textbf{SceneNet RGB-D} dataset provides perfect ground truth annotations for 5M
images. We use a subset: our training set consists of 110,000
images by sampling every 30th frame of each sequence from the first
11 original training splits. Our test dataset consists of 3,000 images by sampling every 100th frame from the original validation set.

All input images are resized to 256
$\times$192. During training, we use data augmentation including
random horizontal flipping and jittering of  brightness
and contrast. At test time, only single scale semantic prediction is evaluated.

\begin{figure*}[!ht]
\centering
\includegraphics[width = 1.8\columnwidth]{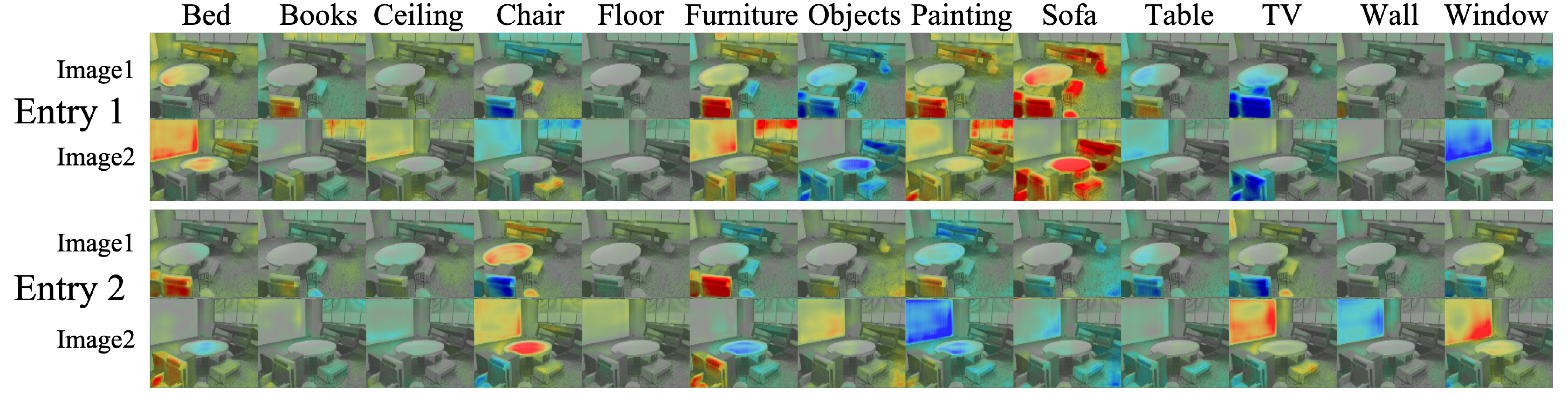}
\caption{The Jacobians of semantic logits w.r.t. two
code entries for a pair of wide baseline views. The columns represent the influence
of the code entry across each semantic classes. Red and
blue mean positive and negative influence, respectively. 
Semantically meaningful regions can be refined coherently during optimisation, leading to smooth and complete segmentation, and this property is automatically carried over into the semantic fusion process.
\label{fig:label_Jacobians}}
\end{figure*}

\subsection{Image Conditioned Scene Representation}
We first quantitively inspect the influence of code size
on the NYUv2 and SceneNet RGB-D datasets by measuring reconstruction
performance. We use the same latent code size for depth
images and semantic labels for simplicity.
We also train a discriminative RefineNet for semantic segmentation
and depth estimation separately as a single task prediction-only baseline models
(i.e.\ code size of 0). Figure \ref{fig:The-reconstruction-accuracy}
shows results for depth and semantic encoding with different code size and setups. The reconstruction performance indicates the capacity of the latent encoding for variational auto-encoders.
Due to the encoded information, the reconstruction
is consistently better than single view monocular prediction.
Furthermore, we do not benefit from a non-linear decoder and we observe diminishing returns when the code size
is larger than 32, and therefore choose this for later experiments.

The qualitative effects of our image conditioned auto-encoding of size 32 are shown in Figure \ref{fig:Qualitative-results}. The zero-code predictions are usually similar to the encoded predictions, though errors in ambiguous regions are corrected given the additional encoded information. Figure \ref{fig:label_Jacobians} displays the learned image dependent Jacobians of the semantic logits w.r.t.\ entries in the code. We see how each code entry is responsible for certain semantically meaningful regions (e.g.\ examine the sofa Jacobians).
Additionally, each code entry also has a tendency to decrease the probability of other ambiguous classes. For two images from different viewpoints, the image dependent Jacobians show high consistency. 
\vspace{-1.5mm}
\subsection{Semantic Label Fusion using Learned Codes}
\begin{figure}[!t]
 \small
 \rotatebox{90}
 {
 \hspace{-2mm} Opt. Entropy
 \hspace{0.5mm} Init. Entropy
 \hspace{2mm} Opt. Label 
  \hspace{3.5mm} Zero code 
 \hspace{5mm} GT Label
 \hspace{4mm} Input image 
 } 
\centering
\includegraphics[height=0.4\paperheight,width=0.95\columnwidth]{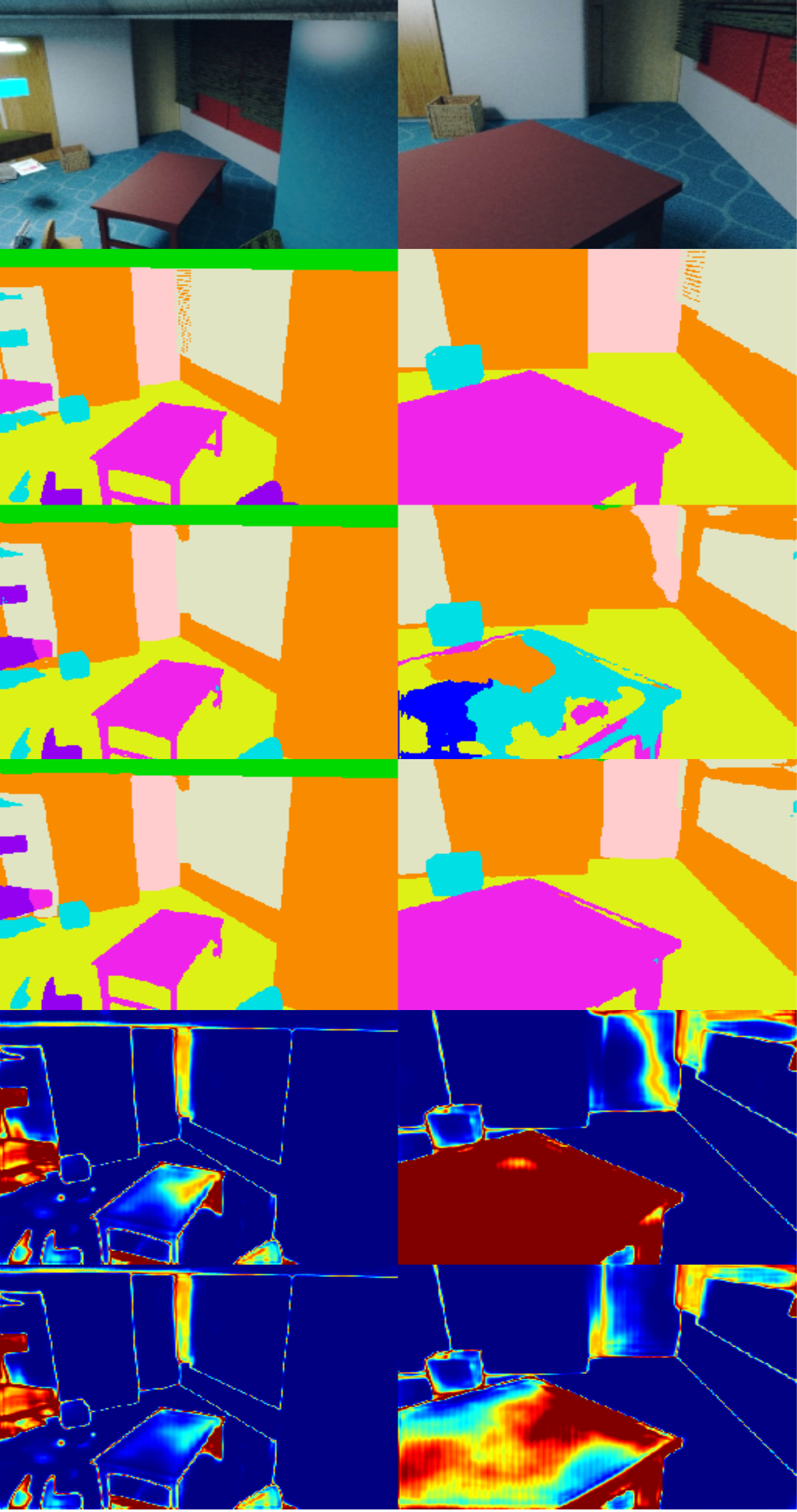}
\caption{An example of two-view semantic label fusion of our method. From top to bottom rows: input colour image, ground truth semantic label, zero-code prediction, optimised label (minimising semantic cost), information entropy of the zero-code softmax probabilities, information entropy of the optimised softmax probabilities. \label{fig:eg_two_view_fusion_table}
}
\end{figure}
\begin{table}[!ht]
\small
\begin{centering}
\begin{tabular}{|c|c|c|c|c|}
\hline 
Statistics & Mean & Std & Max & Min\tabularnewline
\hline 
Rotation (degree) & 5.950 & 9.982 & 163.382 & 0.028\tabularnewline
\hline 
Translation (meter) & 0.149 & 0.087 & 0.701 & 0.001\tabularnewline
\hline 
\end{tabular}
\end{centering}
\caption{The statistics of the relative 3D motion between consecutive frames extracted from SceneNet RGB-D. \label{tab:statistic_eval_sequence} }
\end{table}

Our semantic refinement process can be regarded as a label
fusion scheme for multi-view semantic mapping. An important advantage of code based fusion compared to the usual
element-wise update methods for label fusion is its ability to naturally obtain spatially and temporally consistent
semantic labels by performing joint estimation in the latent space.
This means that pixels are not assumed i.i.d when their semantic probabilities
are updated, leading to smoother and more complete label regions. 

To isolate only label estimation, our experiments use the SceneNet RGB-D dataset where precise ground truth depth and camera poses are available to enable perfect data association. We also mask out and ignore occluded regions. We use the zero-code monocular predictions as the initial semantic predictions for all fusion methods. 

In Figure \ref{fig:eg_two_view_fusion_table} we show the result of semantic label fusion given two views taken with a large baseline. The RHS zero-code prediction struggles to recognise the table given the ambiguous context. The high entropy indicates that  semantic labels are likely to change during optimisation. In contrast, the LHS zero-code prediction is able to accurately segment the table with relatively low entropy. By minimising the semantic cost between two views, the optimised semantic representations are able to generate a consistent predictions, successfully leading to the disambiguation of the RHS into a well segmented and smooth prediction. The entropy of both views are reduced as well. Similar improvements can also be observed in other regions.
In addition, it is interesting to observe that the entropy map exhibits consistency with the scene structure, showing that the network can recognise the spatial extent of an object but struggles with the precise semantic class.

\begin{figure}
\tiny
 \rotatebox{90}
 {
 \hspace{0mm} Opt. Entropy
 \hspace{1.5mm} Init. Entropy
 \hspace{2.5mm} Opt. Label 
 \hspace{4mm} Average
 \hspace{3.5mm} Multiplication
 \hspace{2mm} Zero code
 \hspace{3.5mm} GT Label
 \hspace{4mm} Input image
 } 
\centering
\includegraphics[width=0.95\columnwidth]{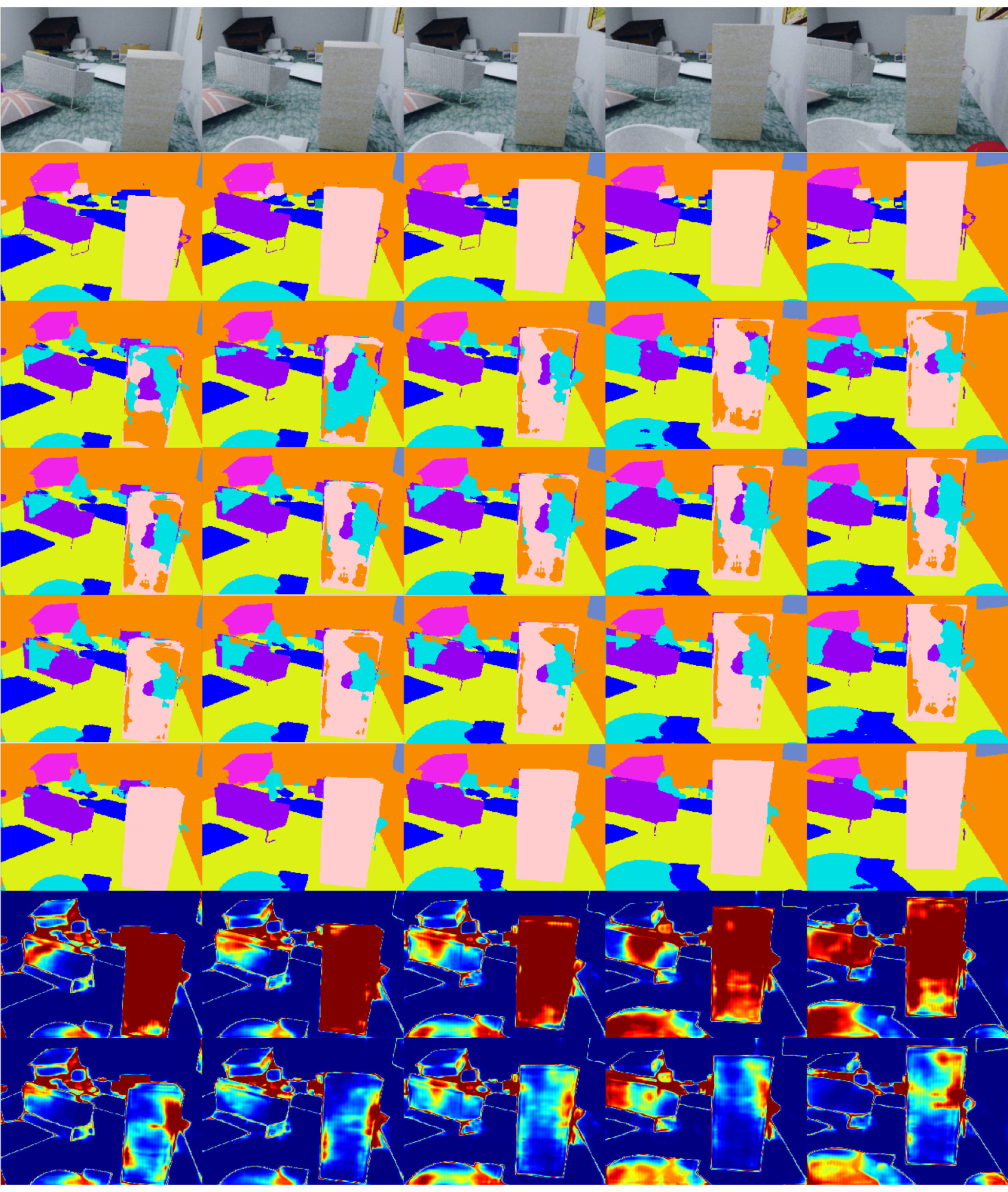}
\caption{Qualitative comparison of different label fusion methods. 5 consecutive frames with a small baseline are chosen. Our method can effectively fuse multi-view semantic labels to generate smoother semantic predictions.
\label{fig:Qualitative-results-of_fusion}}
\vspace{-5mm}
\end{figure}

Qualitative results of different label fusion methods are shown in Figure \ref{fig:Qualitative-results-of_fusion}. The results of both element-wise fusion approaches are obtained by integrating the probabilities of the other images into each current frame, while our result simply comes from pairing all the later frames to the first frame. For a sequence of 5 frames with small baselines, the zero-code predictions are all similar. As a result, when there is a difficult, ambiguous region (indicated by low quality zero-code predictions and high entropy), the element-wise label fusion methods lead to results which are only marginally better. However, the representation power in the learned compact code enables much smoother predictions with correct labels to be obtained through optimisation. After optimisation, the reduced entropy for these regions indicates that the network is much more confident.

Next we provide a quantitative comparison. 2000 images sampled from 1000 sequences (2 images per sequence) from the SceneNet RGB-D validation set are used to evaluate the performance.
We augment every extracted image with a variable number of subsequent images in the sequence to obtain short multi-view sequences (1-4 frames). Since the trajectories of SceneNet RGB-D are randomly generated, a good variety of relative transformations and baselines are included in this set (Table \ref{tab:statistic_eval_sequence}).

Table \ref{tab:The-Label_fusion} shows the effectiveness of three
multi-view label fusion methods given various number of views. Our label fusion approach using code optimisation outperforms others methods. The improvement in total pixel accuracy is not significant because of the large area of walls and floors in the dataset. However, the large improvement in the mIoU metric shows that our method is able to consider more on high-order statistics, indicating smoother predictions and better results on other small objects.

\subsubsection*{Effect of Code Prior during Semantic Optimisation}
During semantic optimisation we use a zero-code regularisation
term. Without this term, the optimisation may be drawn to
locally consistent but incorrect semantic labels. 
Table \ref{tab:The-Label_fusion} shows that the
accuracy of two-view label fusion without a zero-code prior is even lower than single view prediction, underlining the importance of 
this prior.

\begin{table}
\begin{centering}
\small
\begin{tabular}{|c|c|c|c|c|}
\hline 
{\small{}\#Views} & {\small{}Method} & {\small{}Pix. Acc.} & {\small{}Cls. Acc.} & {\small{}mIoU}\tabularnewline
\hline 
{\small{}1} & {\small{}-} & {\small{}75.167} & {\small{}63.330} & {\small{}41.713}\tabularnewline
\hline 
\multirow{4}{*}{{\small{}2}} & {\small{}Multiplication} & {\small{}75.424} & {\small{}63.629} & {\small{}42.326}\tabularnewline
\cline{2-5} 
 & {\small{}Average} & {\small{}75.374} & {\small{}63.549} & {\small{}42.220}\tabularnewline
\cline{2-5} 
 & {\small{}Ours} & \textbf{\small{}75.725} & \textbf{\small{}63.750} & \textbf{\small{}43.842}\tabularnewline
\cline{2-5} 
 & {\small{}Ours \footnotesize{(w/o prior)}} & {\small{}74.498} & {\small{}60.646} & {\small{}39.600}\tabularnewline
\hline 
\multirow{3}{*}{{\small{}3}} & {\small{}Multiplication} & {\small{}75.542} & {\small{}63.815} & {\small{}42.692}\tabularnewline
\cline{2-5} 
 & {\small{}Average} & {\small{}75.451} & {\small{}63.754} & {\small{}42.213}\tabularnewline
\cline{2-5} 
 & {\small{}Ours} & \textbf{\small{}75.815} & \textbf{\small{}63.827} & \textbf{\small{}44.231}\tabularnewline
\hline 
\multirow{3}{*}{{\small{}4}} & {\small{}Multiplication} & {\small{}75.578} & \textbf{\small{}63.950} & {\small{}42.795}\tabularnewline
\cline{2-5} 
 & {\small{}Average} & {\small{}75.358} & {\small{}63.767} & {\small{}42.102}\tabularnewline
\cline{2-5} 
 & {\small{}Ours} & \textbf{\small{}75.668} & {\small{}63.720} & \textbf{\small{}44.263}\tabularnewline
\hline 
\end{tabular}
\end{centering}
\caption{The effectiveness of different label fusion methods on 2000 images sampled from SceneNet RGB-D. The large improvement on the metric of intersection over union shows that our label fusion lead to smoother predictions.
\label{tab:The-Label_fusion} }
\vspace{-4mm}
\end{table}

\begin{figure}[!t]
\centering
\includegraphics[width=0.9\columnwidth]{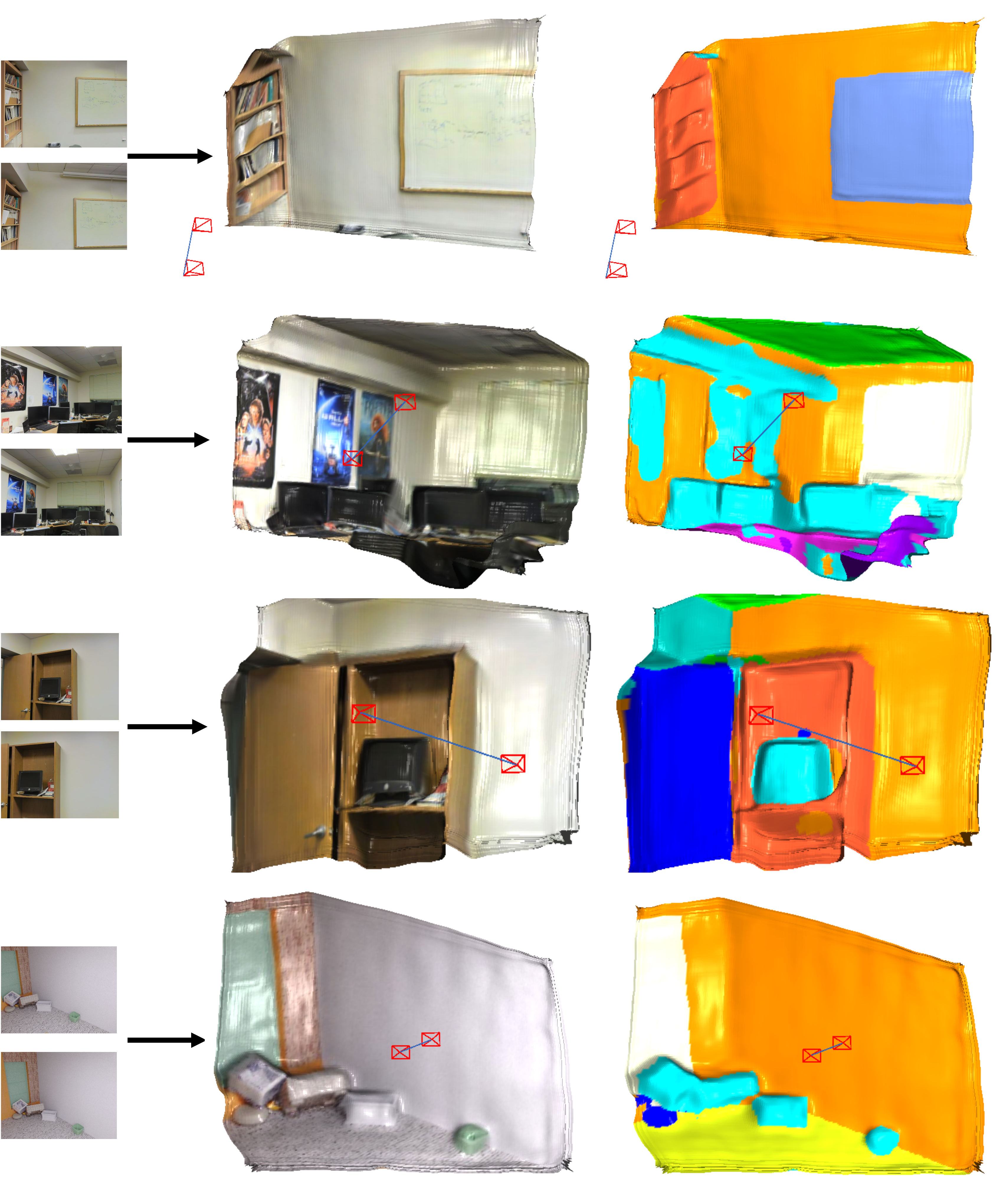}
\caption{Qualitative results of two-view structure from motion on two selected frames from Stanford dataset (first 3 rows) and SceneNet RGB-D dataset (last row). The compact representations of both semantics and geometry are (jointly) optimised with came pose to obtain a dense map with consistent semantic labels and relative camera motion.
\label{fig:Qualitative_SfM}}
\vspace{-7mm}
\end{figure}

\subsection{Monocular Dense Semantic SLAM}
We present example results from our preliminary full monocular dense semantic SLAM system. Due to the prior information of geometry encoded in the system, the system is very robust during initialisation and can manage pure rotational motion. The system currently runs in a sliding window manner. Figures \ref{fig:2view_SFM_bedroom} and \ref{fig:Qualitative_SfM} show examples of two-view dense semantic structure from motion from different datasets.
\vspace{-2mm}
\section{Conclusion and Future Work}
We have shown that an image-conditioned learned compact  representation can coherently and efficiently represent semantic labels.
This code can be optimised across multiple 
overlapping views to implement semantic fusion with many advantages over the usual methods which operate in a per-surface-element independent manner. As well as proving this fusion capability experimentally, we have built and demonstrated a prototype full dense, semantic monocular SLAM system based on learned codes where geometry, poses and semantics can all be jointly optimised.

In future work, we aim to unify learned geometric and semantic representations still further as we continue to reach towards scene models with optimal representational efficiency for truly useful real-time Spatial AI systems.

\section{Acknowledgements}
Research presented in this paper has been supported by Dyson Technology Ltd.
Shuaifeng Zhi holds a China Scholarship Council-Imperial Scholarship. 
We are very grateful to Jan Czarnowski for research and software collaboration on this project.
{\small
\bibliographystyle{ieee_fullname}
\bibliography{SceneCode}

\begin{thebibliography}{10}\itemsep=-1pt

\bibitem{Armeni:etal:ARXIV2017}
Iro {Armeni}, Alexander {Sax}, Amir {R. Zamir}, and Silvio {Savarese}.
\newblock {Joint 2D-3D-Semantic Data for Indoor Scene Understanding}.
\newblock {\em arXiv preprint arXiv:1702.01105}, 2017.

\bibitem{Baker:Matthews:IJCV2004}
Simon Baker and Iain Matthews.
\newblock {{Lucas-Kanade} 20 years on: A Unifying Framework: Part 1}.
\newblock {\em {International Journal of Computer Vision ({IJCV})}},
  56(3):221--255, 2004.

\bibitem{Bloesch:etal:CVPR2018}
Michael Bloesch, Jan Czarnowski, Ronald Clark, Stefan Leutenegger, and
  Andrew~J. Davison.
\newblock {CodeSLAM --- Learning a Compact, Optimisable Representation for
  Dense Visual SLAM}.
\newblock In {\em {Proceedings of the {IEEE} Conference on Computer Vision and
  Pattern Recognition ({CVPR})}}, 2018.

\bibitem{Bowman:etal:ARVIX2015}
Samuel~R. Bowman, Luke Vilnis, Oriol Vinyals, Andrew~M Dai, Rafal Jozefowicz,
  and Samy Bengio.
\newblock {Generating Sentences from a Continuous Space}.
\newblock {\em arXiv preprint arXiv:1511.06349}, 2015.

\bibitem{Cadena:etal:TRO2016}
Cesar Cadena, Luca Carlone, Henry Carrillo, Yasir Latif, Davide Scaramuzza,
  Jos{\'e} Neira, Ian Reid, and John.~J. Leonard.
\newblock {Past, Present, and Future of Simultaneous Localization and Mapping:
  Toward the Robust-Perception Age}.
\newblock {\em {{IEEE} Transactions on Robotics ({T-RO})}}, 32(6):1309--1332,
  Dec 2016.

\bibitem{Cadena:etal:RSS2016}
Cesar Cadena, Anthony~R. Dick, and Ian Reid.
\newblock {Multi-modal Auto-Encoders as Joint Estimators for Robotics Scene
  Understanding}.
\newblock In {\em {Proceedings of Robotics: Science and Systems ({RSS})}},
  2016.

\bibitem{Cha:Srihari:PR2002}
Sung-Hyuk Cha and Sargur~N. Srihari.
\newblock {On Measuring the Distance between Histograms}.
\newblock {\em {Pattern Recognition}}, 35(6):1355--1370, 2002.

\bibitem{Cordts:etal:CVPR2016}
Marius Cordts, Mohamed Omran, Sebastian Ramos, Timo Rehfeld, Markus Enzweiler,
  Rodrigo Benenson, Uwe Franke, Stefan Roth, and Bernt Schiele.
\newblock {The Cityscapes Dataset for Semantic Urban Scene Understanding}.
\newblock In {\em {Proceedings of the {IEEE} Conference on Computer Vision and
  Pattern Recognition ({CVPR})}}, 2016.

\bibitem{Davison:ARXIV2018}
Andrew~J. Davison.
\newblock Futuremapping: {T}he {C}omputational {S}tructure of {S}patial {AI}
  {S}ystems.
\newblock {\em arXiv preprint arXiv:1803.11288}, 2018.

\bibitem{Geiger:etal:CVPR2012}
Andreas Geiger, Philip Lenz, and Raquel Urtasun.
\newblock {Are we ready for Autonomous Driving? The KITTI Vision Benchmark
  Suite}.
\newblock In {\em {Proceedings of the {IEEE} Conference on Computer Vision and
  Pattern Recognition ({CVPR})}}, 2012.

\bibitem{He:etal:ICCV2015}
Kaiming He, Xiangyu Zhang, Shaoqing Ren, and Jian Sun.
\newblock {Delving Deep into Rectifiers: Surpassing Human-Level Performance on
  ImageNet Classification}.
\newblock In {\em {Proceedings of the International Conference on Computer
  Vision ({ICCV})}}, 2015.

\bibitem{He:etal:CVPR2016}
Kaiming He, Xiangyu Zhang, Shaoqing Ren, and Jian Sun.
\newblock {Deep Residual Learning for Image Recognition}.
\newblock In {\em {Proceedings of the {IEEE} Conference on Computer Vision and
  Pattern Recognition ({CVPR})}}, 2016.

\bibitem{He:etal:CVPR2017}
Yang He, Wei-Chen Chiu, Margret Keuper, and Mario Fritz.
\newblock {STD2P: RGBD Semantic Segmentation Using Spatio-Temporal Data-Driven
  Pooling}.
\newblock In {\em {Proceedings of the {IEEE} Conference on Computer Vision and
  Pattern Recognition ({CVPR})}}, 2017.

\bibitem{Hermans:etal:ICRA2014}
Alexander Hermans, Georgios Floros, and Bastian Leibe.
\newblock {Dense 3D Semantic Mapping of Indoor Scenes from RGB-D Images}.
\newblock In {\em {Proceedings of the {IEEE} International Conference on
  Robotics and Automation ({ICRA})}}, 2014.

\bibitem{Isola:etal:CVPR2017}
Phillip Isola, Jun-Yan Zhu, Tinghui Zhou, and Alexei~A. Efros.
\newblock Image-to-image translation with conditional adversarial networks.
\newblock In {\em {Proceedings of the {IEEE} Conference on Computer Vision and
  Pattern Recognition ({CVPR})}}, 2017.

\bibitem{Kahler:Reid:ICCV2013}
Olaf Kahler and Ian Reid.
\newblock {Efficient 3D Scene Labelling Using Fields of Trees}.
\newblock In {\em Proceedings of the IEEE International Conference on Computer
  Vision}, 2013.

\bibitem{Kendall:Gal:NIPS2017}
Alex Kendall and Yarin Gal.
\newblock {What Uncertainties Do We Need in Bayesian Deep Learning for Computer
  Vision?}
\newblock In {\em {Neural Information Processing Systems ({NIPS})}}, 2017.

\bibitem{Kendall:etal:CVPR2018}
Alex Kendall, Yarin Gal, and Roberto Cipolla.
\newblock Multi-{T}ask {L}earning {U}sing {U}ncertainty to {W}eigh {L}osses for
  {S}cene {G}eometry and {S}emantics.
\newblock In {\em {Proceedings of the {IEEE} Conference on Computer Vision and
  Pattern Recognition ({CVPR})}}, 2018.

\bibitem{Kerl:etal:ICRA2013}
Christian Kerl, J{\"u}rgen Sturm, and Daniel Cremers.
\newblock {Robust Odometry Estimation for {RGB-D} Cameras}.
\newblock In {\em {Proceedings of the {IEEE} International Conference on
  Robotics and Automation ({ICRA})}}, 2013.

\bibitem{Kingma:Ba:ICLR2015}
Diederik~P. Kingma and Jimmy Ba.
\newblock {Adam: {A} Method for Stochastic Optimization}.
\newblock In {\em {Proceedings of the International Conference on Learning
  Representations ({ICLR})}}, 2015.

\bibitem{Kingma:Welling:ICLR2014}
Diederik~P. Kingma and Max Welling.
\newblock {Auto-Encoding Variational Bayes}.
\newblock In {\em {Proceedings of the International Conference on Learning
  Representations ({ICLR})}}, 2014.

\bibitem{Klein:Murray:ISMAR2007}
Georg Klein and David~W. Murray.
\newblock {Parallel Tracking and Mapping for Small {AR} Workspaces}.
\newblock In {\em {Proceedings of the International Symposium on Mixed and
  Augmented Reality ({ISMAR})}}, 2007.

\bibitem{Kohl:etal:NIPS2018}
Simon A.~A. Kohl, Bernardino Romera-Paredes, Clemens Meyer, Jeffrey De~Fauw,
  Joseph~R. Ledsam, Klaus~H. Maier-Hein, S.~M. Eslami, Danilo~Jimenez Rezende,
  and Olaf Ronneberger.
\newblock A {P}robabilistic {U}-{N}et for {S}egmentation of {A}mbiguous
  {I}mages.
\newblock In {\em {Neural Information Processing Systems ({NIPS})}}, 2018.

\bibitem{Lin:etal:CVPR2017}
Guosheng Lin, Anton Milan, Chunhua Shen, and Ian Reid.
\newblock {RefineNet: Multi-Path Refinement Networks for High-Resolution
  Semantic Segmentation}.
\newblock In {\em {Proceedings of the {IEEE} Conference on Computer Vision and
  Pattern Recognition ({CVPR})}}, 2017.

\bibitem{Long:etal:CVPR2015}
Jonathan Long, Evan Shelhamer, and Trevor Darrell.
\newblock {Fully Convolutional Networks for Semantic Segmentation}.
\newblock In {\em {Proceedings of the {IEEE} Conference on Computer Vision and
  Pattern Recognition ({CVPR})}}, 2015.

\bibitem{Ma:etal:IROS2017}
Lingni Ma, J{\"o}rg St{\"u}ckler, Christian Kerl, and Daniel Cremers.
\newblock {Multi-View Deep Learning for Consistent Semantic Mapping with RGB-D
  Cameras}.
\newblock In {\em {Proceedings of the {IEEE/RSJ} Conference on Intelligent
  Robots and Systems ({IROS})}}, 2017.

\bibitem{Mccormac:etal:3DV2018}
John McCormac, Ronald Clark, Michael Bloesch, Andrew~J. Davison, and Stefan
  Leutenegger.
\newblock {{Fusion\texttt{++}}:Volumetric Object-Level SLAM}.
\newblock In {\em {Proceedings of the International Conference on 3D Vision
  ({3DV})}}, 2018.

\bibitem{McCormac:etal:ICRA2017}
John McCormac, Ankur Handa, Andrew~J. Davison, and Stefan Leutenegger.
\newblock {{SemanticFusion}: Dense {3D} Semantic Mapping with Convolutional
  Neural Networks}.
\newblock In {\em {Proceedings of the {IEEE} International Conference on
  Robotics and Automation ({ICRA})}}, 2017.

\bibitem{McCormac:etal:ICCV2017}
John McCormac, Ankur Handa, Stefan Leutenegger, and Andrew~J. Davison.
\newblock {{SceneNet RGB-D}: Can {5M} Synthetic Images Beat Generic {ImageNet}
  Pre-training on Indoor Segmentation?}
\newblock In {\em {Proceedings of the International Conference on Computer
  Vision ({ICCV})}}, 2017.

\bibitem{Nicholson:etal:RAL2018}
Lachlan Nicholson, Michael Milford, and Niko S\"{u}nderhauf.
\newblock {QuadricSLAM: Constrained Dual Quadrics from Object Detections as
  Landmarks in Object-oriented SLAM}.
\newblock {\em {{IEEE} Robotics and Automation Letters}}, 2018.

\bibitem{Runz:etal:ISMAR2018}
Martin R{\"u}nz, Maud Buffier, and Lourdes Agapito.
\newblock {MaskFusion: Real-time Recognition, Tracking and Reconstruction of
  Multiple Moving Objects}.
\newblock In {\em {Proceedings of the International Symposium on Mixed and
  Augmented Reality ({ISMAR})}}, 2018.

\bibitem{Silberman:etal:ECCV2012}
Nathan Silberman, Derek Hoiem, Pushmeet Kohli, and Rob Fergus.
\newblock {Indoor Segmentation and Support Inference from RGBD Images}.
\newblock In {\em {Proceedings of the European Conference on Computer Vision
  ({ECCV})}}, 2012.

\bibitem{Sohn:etal:NIPS2015}
Kihyuk Sohn, Honglak Lee, and Xinchen Yan.
\newblock {Learning Structured Output Representation using Deep Conditional
  Generative Models}.
\newblock In {\em {Neural Information Processing Systems ({NIPS})}}, 2015.

\bibitem{Sonderby:etal:arXiv2016}
Casper~Kaae S{\o}nderby, Tapani Raiko, Lars Maal{\o}e, S{\o}ren~Kaae
  S{\o}nderby, and Ole Winther.
\newblock {How to Train Deep Variational Autoencoders and Probabilistic Ladder
  Networks}.
\newblock {\em arXiv preprint arXiv:1602.02282}, 2016.

\bibitem{Sunderhauf:etal:IROS2017}
Niko S\"{u}nderhauf, Trung~T. Pham, Yasir Latif, Michael Milford, and Ian Reid.
\newblock {Meaningful Maps With Object-Oriented Semantic Mapping}.
\newblock In {\em {Proceedings of the {IEEE/RSJ} Conference on Intelligent
  Robots and Systems ({IROS})}}, 2017.

\bibitem{Weerasekera:etal:ICRA2017}
Chamara~Saroj Weerasekera, Yasir Latif, Ravi Garg, and Ian Reid.
\newblock {Dense Monocular Reconstruction using Surface Normals}.
\newblock In {\em {Proceedings of the {IEEE} International Conference on
  Robotics and Automation ({ICRA})}}, 2017.

\bibitem{Jiajun:etal:NIPS2015}
Jiajun Wu, Ilker Yildirim, Joseph~J. Lim, William~T. Freeman, and Joshua~B.
  Tenenbaum.
\newblock {Galileo: Perceiving Physical Object Properties by Integrating a
  Physics Engine with Deep Learning}.
\newblock In {\em {Neural Information Processing Systems ({NIPS})}}, 2015.

\bibitem{Xiang:Fox:ARXIV2017}
Yu Xiang and Dieter Fox.
\newblock {{DA-RNN}: Semantic Mapping with Data Associated Recurrent Neural
  Networks}.
\newblock {\em arXiv preprint arXiv:1703.03098}, 2017.

\bibitem{Xiao:Quan:ICCV2009}
Jianxiong Xiao and Long Quan.
\newblock {Multiple View Semantic Segmentation for Street View Images}.
\newblock In {\em {Proceedings of the International Conference on Computer
  Vision ({ICCV})}}, 2009.

\end{thebibliography}
}
\end{document}